\documentclass[letterpaper, 10 pt, conference]{ieeeconf}  

\IEEEoverridecommandlockouts                              
\usepackage{xcolor}
\usepackage[inkscapelatex=false]{svg}
\usepackage{comment}
\usepackage{graphics} 
\usepackage{epsfig} 
\usepackage{amsmath} 
\usepackage{amssymb}  
\usepackage{mathtools}
\usepackage{marginnote}
\usepackage{hyperref}
\usepackage{cite}
\usepackage{mathrsfs}
\usepackage{bm}
\usepackage{amsfonts}
\usepackage{booktabs}
\usepackage{tabularx}
\usepackage[nameinlink,capitalize]{cleveref}
\crefname{equation}{Eq.}{Eqs.}
\Crefname{equation}{Eq.}{Eqs.}
\crefname{algorithm}{Alg.}{Algs.}
\Crefname{algorithm}{Alg.}{Algs.}
\crefname{section}{Sec.}{Secs.}
\Crefname{section}{Sec.}{Secs.}
\title{\LARGE \bf
Simultaneous Calibration of Noise Covariance and Kinematics for State Estimation of Legged Robots via Bi-level Optimization} 

\author{
Denglin Cheng$^{1,*}$, Jiarong Kang$^{2,*}$, and Xiaobin Xiong$^{3}$
\thanks{Code is available at \url{https://github.com/DLinC3/LegBiCal}}
\thanks{$^{*}$Equal contribution.}%
\thanks{All authors are with the Legged AI Lab.}%
\thanks{$^{1}$D. Cheng is now with the LCSR, Johns Hopkins University, MD, USA.}%
\thanks{$^{2}$J. Kang is now with the University of Wisconsin--Madison, WI, USA.}%
\thanks{$^{3}$X. Xiong is now with the Shanghai Innovation Institute (SII), Shanghai, China,
and was with the University of Wisconsin--Madison.
Corresponding to X. Xiong (\tt\small xiaobin.xiong@sii.edu.cn)}%
}
\begin{document}

\newcommand{\TODO}[1]{{\color{blue} #1}}
\newcommand{\fix}[1]{{\color{black} #1}}

 \newcommand{\block}[1]{\noindent{\textbf{#1}:}}
\newcommand{\emphhh}[1]{{\color{yellow} \textbf{#1}}}

\newcommand{\newhline}{\ \hline \ }
\maketitle
\thispagestyle{empty}
\pagestyle{empty}

\begin{abstract}
Accurate state estimation is critical for legged and aerial robots operating in dynamic, uncertain environments. A key challenge lies in specifying process and measurement noise covariances, which are typically unknown or manually tuned. In this work, we introduce a bi-level optimization framework that jointly calibrates covariance matrices and kinematic parameters in an estimator-in-the-loop manner. The upper level treats noise covariances and model parameters as optimization variables, while the lower level executes a full-information estimator. Differentiating through the estimator allows direct optimization of trajectory-level objectives, resulting in accurate and consistent state estimates. We validate our approach on quadrupedal and humanoid robots, demonstrating significantly improved estimation accuracy and uncertainty calibration compared to hand-tuned baselines. Our method unifies state estimation, sensor, and kinematics calibration into a principled, data-driven framework applicable across diverse robotic platforms. Video available at: \url{https://youtu.be/1zFORUMdLbg}.
\end{abstract}

\section{Introduction}
State estimation is fundamental to the autonomy of robotic systems, providing the basis for planning and control. Robots must infer their states--such as position, velocity, and orientation-from noisy sensor data, often under challenging real-world conditions. Classical approaches such as the Kalman filter and its extensions \cite{kalman1960, bai2023state} remain central, and have been successfully deployed in applications ranging from ground vehicles \cite{thrun2005probabilistic} and aerial robots \cite{mohamed2021momentum} to quadrupeds and humanoids \cite{bloesch2012state, rotella2014state, he2025invariant, HLIP}. More recently, factor graph–based methods and smoothing techniques \cite{kaess2012isam2, dellaert2017factor} have enabled large-scale estimation for simultaneous localization and mapping (SLAM) \cite{cadena2017past} and visual–inertial odometry \cite{mur2017orb,usenko2019visual}.

A persistent challenge in these methods is the specification of noise covariance matrices. The process noise covariance captures uncertainty in the dynamics and actuation, while the measurement noise covariance characterizes sensor error models. In practice, both are difficult to obtain: manufacturer datasheets provide only partial information, while individual sensor calibration and physical system identification are both expensive and task-dependent. As a result, practitioners often rely on manual tuning \cite{he2025invariant, hartley2020contact, kang24} - a heuristic, time-consuming process that can produce estimators that are either sub-optimal or inconsistent \cite{thrun2005probabilistic} over out-of-distribution tasks. This problem is particularly acute for highly dynamic robots such as quadrupeds and humanoids, whose complex, high-dimensional hybrid dynamics and kinematics, and multi-modal sensor nonlinearities amplify the challenges to obtain precise state estimates \cite{rotella2014state, hartley2020contact, he2025invariant, HLIP, kang24}, especially under kinematic uncertainties.
Several lines of work have been proposed to address this issue. Adaptive filtering methods \cite{9044358} adjust covariances online based on innovation statistics, while expectation–maximization (EM) frameworks \cite{em_mhe_2017, 10472604, yap2008simultaneous} iteratively refine noise parameters to maximize likelihood \cite{6630924, 6482657}. Learning sensor parameters or noises parameters \cite{wong2020variational, qadri2024learning, yap2008simultaneous} is another common approach. However, these methods often either assume no availability of ground truth data or only focus on decoupled calibrations. More importantly, their treatments have not been examined on modern legged robots that have multimodal sensor suites and complex dynamics.

\begin{figure}[t]
    \centering
    \includesvg[width=0.8\linewidth]{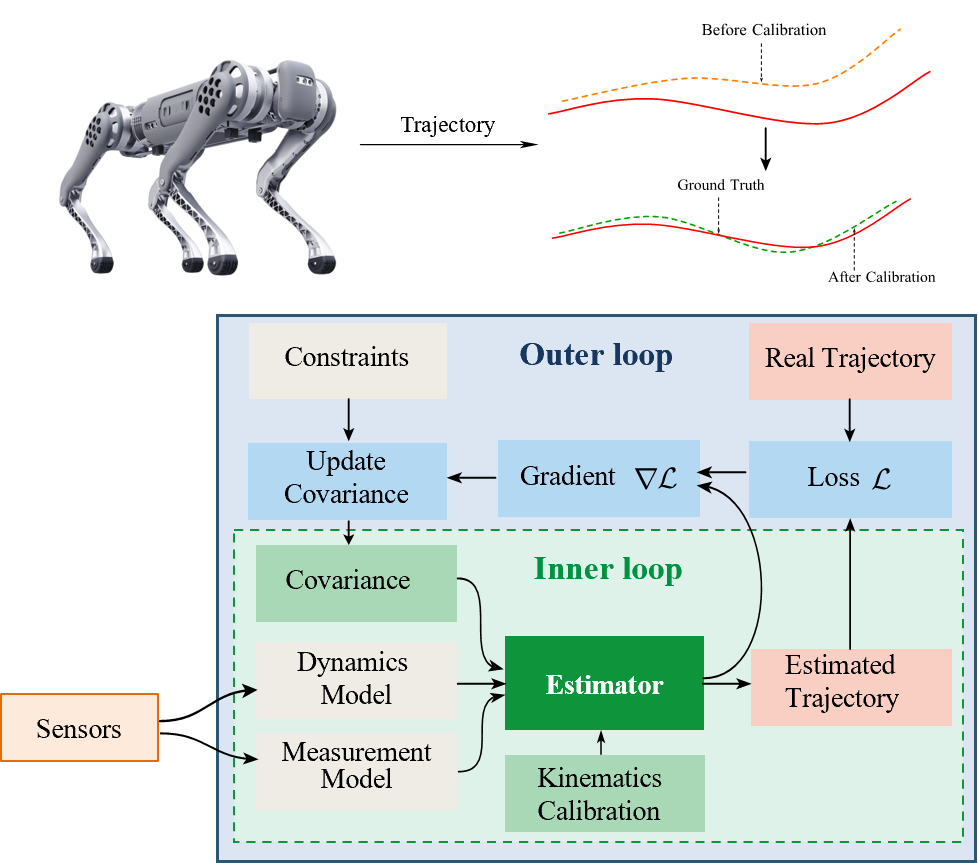}
    \caption{Overview of the work with its application to a quadrupedal robot. }
    \label{fig:overview}
\end{figure}

In this work, we propose a bi-level optimization framework for simultaneous covariance and kinematics calibration for robotic state estimation. Figure~\ref{fig:overview} summarizes the estimator-in-the-loop construction. At the upper level, we treat covariance matrices and uncertain kinematics as optimization variables. At the lower level, we solve a full-information estimator using the Maximum a Posteriori (MAP) formulation given the current covariance estimates and kinematics information. By differentiating through the estimator \cite{amos2017optnet,gould2016differentiating}, we directly optimize parameters to be calibrated with respect to trajectory-level objectives of minimizing the errors between the estimates and the measurable ground truth. This approach eliminates manual tuning, enforces physical consistency through structured parameterization and constraints, and generalizes across sensor modalities and robots.

We validate our approach on quadrupedal and bipedal robots, comparing against hand-tuned baselines. Across both platforms, our method calibrate the parameters to physically plausible values and significantly reduces estimation error and improves uncertainty consistency. These results highlight bi-level optimization for calibration as a principled and practical tool for bridging the gap between theoretical estimation frameworks and the demands of real-world robotics.
 
\section{Related Work}
 
\block{Legged Robotic State Estimation} State estimation for legged robots has been extensively studied due to the challenges posed by intermittent ground contacts, actuator uncertainties, and highly dynamic motions. Early approaches utilize Extended Kalman Filters (EKFs) to fuse proprioceptive sensing - Inertial Measurement Unit (IMU), contact sensors, and joint encoders - with kinematic constraints to achieve drift-reduced estimates on quadrupedal \cite{bloesch2012state}, bipedal \cite{teng2021legged} and humanoid robots \cite{rotella2014state}. Recent focuses of using EKFs on legged robots employ the invariant properties of the SE(3) manifold, and thus yield more accurate estimates \cite{he2025invariant, hartley2020contact}. Another line of work leveraged factor graphs \cite{dellaert2017factor, yang2024real} and smoothing or windowed formulations such as Moving Horizon Estimation (MHE) \cite{kang24} to integrate proprioceptive and exteroceptive sensing data at scale. Despite these advances, most approaches rely on fixed or manually tuned noise covariances, limiting the accuracy of estimates, especially in the presence of kinematic uncertainty. Our work addresses this gap by introducing a bi-level optimization framework that calibrates covariance matrices and uncertain kinematics simultaneously in a principled manner.

\block{Sensor Calibration} Sensor calibration and noise characterization have also been studied extensively in robotics, since the accuracy of estimation strongly depends on the quality of sensor models. Traditional approaches focus on extrinsic and intrinsic calibration of individual sensors, such as cameras \cite{rehder2016extending}, LiDARs \cite{li2023automatic}, and IMUs \cite{fu2021high}. At the theoretical level, several works address the identification of noise statistics. For instance, expectation–maximization (EM) algorithms iteratively refine noise models to maximize likelihood \cite{em_mhe_2017, 10472604, yap2008simultaneous}. Recent efforts explored data-driven estimation of uncertainty, including machine learning to optimize covariances from raw sensory inputs \cite{qadri2024learning, wang2023neural}. Despite these advances, most existing techniques commonly are tailored to specific sensors rather than complex legged robots; they often assume restrictive statistical models, or require careful initialization. In contrast, our bi-level optimization jointly calibrates both process and measurement noise with uncertain kinematics across diverse and complex legged robotic platforms.

\block{System Identification} System identification (SysID) has long been a cornerstone of robotics, enabling accurate modeling of dynamics, kinematics, and actuation. Classical techniques for parameter estimation of rigid-body kinematics or dynamics exploit regressions of the geometric or Lagrangian models \cite{leboutet2021inertial, 7987066, khorshidi2025physically} subject to physical constraints. Gaussian processes or neural networks approximate residuals not captured by physics-based models \cite{hwangbo2019learning} are commonly used in calibration. Recent works \cite{yang2022online, burgul2024online} focus on online kinematics calibration via state estimators. Despite significant progress, SysID typically treats noise statistics, model parameters, and estimator design separately. Our bi-level optimization approach bridges this gap by jointly calibrating positive definite covariance matrices and kinematic parameters, highlighting the natural synergy between system identification and state estimation.


\section{Modeling}\label{sec:prelim}

We consider two state-estimation models within a common calibration framework.
Section~\ref{sec:mobile} combines inertial propagation with contact-conditioned
leg kinematics, whereas Section~\ref{sec:prime_estimator} uses differentiable
multibody contact dynamics.  Section~\ref{sec:map} casts either model as a
full-information estimation (FIE) problem.

Upright capital letters denote frames: $\mathrm W$ is the world frame,
$\mathrm B$ is fixed to the floating base, and $\mathrm F_j$ is the nominal
foot frame of leg $j$.  The symbol ${}^{A}\mathbf p^{C}_{F}$ denotes the
position of $C$ measured from $A$ and expressed in $F$, and
${}^{A}\mathbf R^{B}$ maps coordinates from $B$ to $A$.  The paired frame
labels distinguish rotation matrices from estimator covariances: throughout,
$\mathbf R$ without such labels denotes a measurement covariance.
Translational quantities lie in $\mathbb R^3$, and each point contact is
resolved into one normal and two tangential components.

\begin{figure}[!htbp]
  \centering
  \includesvg[width=0.7\linewidth]{figure/humanrobot.svg}
  \caption{Common proprioceptive sensing on the Unitree Go1 and G1: an IMU,
  joint encoders, and foot-contact sensing.}
  \label{fig:legged-sensing}
\end{figure}

\subsection{Legged Robot State Estimation}\label{sec:mobile}

We fuse the proprioceptive signals in Fig.~\ref{fig:legged-sensing}, using IMU
measurements and contact-conditioned position and velocity observations from leg kinematics
\cite{bloesch2012state,camurri2020pronto,teng2021legged,yang2024real}.  The state
contains the base pose and velocity, the world position of each foot, and the
IMU biases:
\begin{equation}
  \mathbf x_k:=\bigl(
  {}^{\mathrm W}\mathbf X^{\mathrm B}_k,
  {}^{\mathrm W}\mathbf v^{\mathrm B}_{\mathrm W,k},
  \{{}^{\mathrm W}\mathbf p^{\mathrm F_j}_{\mathrm W,k}\}_{j=1}^{n_f},
  \mathbf b_{a,k},\mathbf b_{\omega,k}\bigr).
  \label{eq:state_def}
\end{equation}
Here ${}^{\mathrm W}\mathbf X^{\mathrm B}_k\in\mathrm{SE}(3)$ is the
floating-base pose and comprises
$({}^{\mathrm W}\mathbf p^{\mathrm B}_{\mathrm W,k},
{}^{\mathrm W}\mathbf R^{\mathrm B}_k)$; base velocity is expressed in
$\mathrm W$, and the biases are expressed in $\mathrm B$.  An external contact
indicator $c_{j,k}\in\{0,1\}$ specifies whether leg $j$ is treated as a
non-slipping stance contact, and
$\mathcal I^{\mathrm{st}}_k:=\{j\mid c_{j,k}=1\}$ denotes the stance set.

\subsubsection{Process Model}\label{sec:process}

With additive white sensor noise, the accelerometer and gyroscope models are
\begin{align*}
  \widetilde{\mathbf a}^{\mathrm B}_k
  &= {}^{\mathrm B}\mathbf R^{\mathrm W}_k
  ({}^{\mathrm W}\dot{\mathbf v}^{\mathrm B}_{\mathrm W,k}
  -{}^{\mathrm W}\mathbf g_{\mathrm W})
  +\mathbf b_{a,k}+\boldsymbol\epsilon_{a,k},\\
  \widetilde{\boldsymbol\omega}^{\mathrm B}_k
  &= {}^{\mathrm W}\boldsymbol\omega^{\mathrm B}_{\mathrm B,k}
  +\mathbf b_{\omega,k}+\boldsymbol\epsilon_{\omega,k}.
\end{align*}
Under a zero-order hold over $[t_k,t_{k+1})$, the inferred world-frame
acceleration is
\begin{equation*}
  \mathbf a^{\mathrm W}_k
  :={}^{\mathrm W}\mathbf R^{\mathrm B}_k
  (\widetilde{\mathbf a}^{\mathrm B}_k-\mathbf b_{a,k})
  +{}^{\mathrm W}\mathbf g_{\mathrm W}.
\end{equation*}
The corresponding nominal discrete process is
\begin{align*}
  {}^{\mathrm W}\mathbf p^{\mathrm B}_{\mathrm W,k+1}
  &= {}^{\mathrm W}\mathbf p^{\mathrm B}_{\mathrm W,k}
  +h{}^{\mathrm W}\mathbf v^{\mathrm B}_{\mathrm W,k}
  +\tfrac12h^2\mathbf a^{\mathrm W}_k,\\
  {}^{\mathrm W}\mathbf v^{\mathrm B}_{\mathrm W,k+1}
  &= {}^{\mathrm W}\mathbf v^{\mathrm B}_{\mathrm W,k}
  +h\mathbf a^{\mathrm W}_k,\\
  {}^{\mathrm W}\mathbf R^{\mathrm B}_{k+1}
  &= {}^{\mathrm W}\mathbf R^{\mathrm B}_k
  \operatorname{Exp}\!\left(
  h[\widetilde{\boldsymbol\omega}^{\mathrm B}_k
  -\mathbf b_{\omega,k}]_\times\right),\\
  {}^{\mathrm W}\mathbf p^{\mathrm F_j}_{\mathrm W,k+1}
  &= {}^{\mathrm W}\mathbf p^{\mathrm F_j}_{\mathrm W,k},
  \qquad
  \mathbf b_{(\cdot),k+1}=\mathbf b_{(\cdot),k}.
\end{align*}
The foot-position process covariance is contact
dependent~\cite{bloesch2012state,yang2024real},
\begin{equation}
  \mathbf Q_{p,j,k}
  =c_{j,k}\mathbf Q^{\mathrm{st}}_{p,j}
  +(1-c_{j,k})\mathbf Q^{\mathrm{sw}}_{p,j},
  \qquad
  \mathbf Q^{\mathrm{sw}}_{p,j}\succeq\mathbf Q^{\mathrm{st}}_{p,j},
  \label{eq:foot_process_covariance}
\end{equation}
The remaining process covariance accounts for IMU noise, bias drift, and
discretization error.

\subsubsection{Measurement Model}\label{sec:measurement}

Let
$\mathbf r_j(\mathbf q_j;\boldsymbol\rho):=
\operatorname{fk}_j(\mathbf q_j;\boldsymbol\rho)$ be the position of
$\mathrm F_j$ relative to $\mathrm B$, expressed in $\mathrm B$, and let
$\mathbf J_j:=D_{\mathbf q_j}\mathbf r_j$.  The relevant components of
$\boldsymbol\rho_{\mathrm{foot}}$ parameterize the terminal-link geometry.
The rigid kinematic identity
${}^{\mathrm W}\mathbf p^{\mathrm F_j}_{\mathrm W}
={}^{\mathrm W}\mathbf p^{\mathrm B}_{\mathrm W}
+{}^{\mathrm W}\mathbf R^{\mathrm B}\mathbf r_j$
gives the position observation
\begin{subequations}\label{eq:meas_models}
\begin{align}
  \mathbf H_{p,j}(\mathbf x_k)
  &= {}^{\mathrm B}\mathbf R^{\mathrm W}_k
  ({}^{\mathrm W}\mathbf p^{\mathrm F_j}_{\mathrm W,k}
  -{}^{\mathrm W}\mathbf p^{\mathrm B}_{\mathrm W,k}),\\
  \widetilde{\mathbf y}_{p,j,k}
  &=\mathbf r_j(\widetilde{\mathbf q}_{j,k};\boldsymbol\rho).
\end{align}
For $j\in\mathcal I^{\mathrm{st}}_k$, differentiating the same identity and imposing
${}^{\mathrm W}\dot{\mathbf p}^{\mathrm F_j}_{\mathrm W}=\mathbf0$ yields
\begin{equation*}
  -{}^{\mathrm B}\mathbf R^{\mathrm W}_k
  {}^{\mathrm W}\mathbf v^{\mathrm B}_{\mathrm W,k}
  =\mathbf J_j\dot{\mathbf q}_{j,k}
  +[{}^{\mathrm W}\boldsymbol\omega^{\mathrm B}_{\mathrm B,k}]_\times
  \mathbf r_j.
\end{equation*}
and hence the velocity observation
\begin{align}
  \mathbf H_{v,j}(\mathbf x_k;\widetilde{\mathbf q}_{j,k})
  &=-{}^{\mathrm B}\mathbf R^{\mathrm W}_k
  {}^{\mathrm W}\mathbf v^{\mathrm B}_{\mathrm W,k}
  +[\mathbf b_{\omega,k}]_\times
  \mathbf r_j(\widetilde{\mathbf q}_{j,k};\boldsymbol\rho),\\
  \widetilde{\mathbf y}_{v,j,k}
  &=\mathbf J_j(\widetilde{\mathbf q}_{j,k};\boldsymbol\rho)
  \dot{\widetilde{\mathbf q}}_{j,k}
  +[\widetilde{\boldsymbol\omega}^{\mathrm B}_k]_\times
  \mathbf r_j(\widetilde{\mathbf q}_{j,k};\boldsymbol\rho).
\end{align}
\end{subequations}
The position residual is retained for every leg, with swing-foot motion
handled through \eqref{eq:foot_process_covariance}; the velocity residual is
used only for $j\in\mathcal I^{\mathrm{st}}_k$.

To propagate encoder and gyroscope uncertainty, define the stacked residual
$\mathbf r^{\mathrm{kin}}_{j,k}:=\operatorname{col}(
\widetilde{\mathbf y}_{p,j,k}-\mathbf H_{p,j},
\widetilde{\mathbf y}_{v,j,k}-\mathbf H_{v,j})$ and the local sensor vector
$\widetilde{\boldsymbol\xi}_{j,k}:=\operatorname{col}(
\widetilde{\mathbf q}_{j,k},
\dot{\widetilde{\mathbf q}}_{j,k},
\widetilde{\boldsymbol\omega}^{\mathrm B}_k)$.  The corresponding residual
Jacobian is
\begin{equation}
\begin{aligned}
\mathbf J_{\xi,j,k}:=D_{\widetilde{\boldsymbol\xi}_{j,k}}
\mathbf r^{\mathrm{kin}}_{j,k}
&=
\begin{bmatrix}
  \mathbf J_j & \mathbf0 & \mathbf0\\
  \mathbf A_{j,k} &\mathbf J_j&-[\mathbf r_j]_\times
\end{bmatrix},\\
\mathbf A_{j,k}&:=D_{\mathbf q_j}\mathbf J_j
[\dot{\widetilde{\mathbf q}}_{j,k}]
+[\widetilde{\boldsymbol\omega}^{c}_k]_\times\mathbf J_j,\\
\widetilde{\boldsymbol\omega}^{c}_k
&:=\widetilde{\boldsymbol\omega}^{\mathrm B}_k-\mathbf b_{\omega,k},
\end{aligned}
\label{eq:kinematic-residual-jacobian}
\end{equation}
where $\mathbf r_j$ and $\mathbf J_j$ are evaluated at
$(\widetilde{\mathbf q}_{j,k};\boldsymbol\rho)$, and
$D_{\mathbf q_j}\mathbf J_j[\mathbf d]$ is the directional derivative along
$\mathbf d$.
When all residuals are stacked, let $\widetilde{\boldsymbol\xi}_k$ contain
each encoder channel and the shared gyroscope measurement once, and let
$\boldsymbol\Sigma_{\xi,k}$ denote its covariance.  The stacked residual
$\mathbf r^{\mathrm{kin}}_k$ has Jacobian
$\mathbf J_{\xi,k}:=D_{\widetilde{\boldsymbol\xi}_k}
\mathbf r^{\mathrm{kin}}_k$, assembled from
\eqref{eq:kinematic-residual-jacobian}.  Its effective covariance is
\begin{equation*}
  \mathbf R^{\mathrm{kin}}_k
  =\mathbf J_{\xi,k}\boldsymbol\Sigma_{\xi,k}\mathbf J_{\xi,k}^{\top}
  +\mathbf R_{\mathrm{add},k}.
\end{equation*}
Here $\mathbf R_{\mathrm{add},k}$ captures residual kinematic and contact-model
error.  The construction retains the position--velocity correlation induced by
joint encoders and the cross-leg correlation induced by the shared gyroscope.
Consequently, $\boldsymbol\rho$ changes both the kinematic residuals and their
first-order covariance.

\subsection{Contact-Aware Modeling}\label{sec:prime_estimator}

\subsubsection{Process Modeling}\label{sec:prime_process}

PRIME~\cite{Kang2026PRIME} embeds a log-barrier smoothing of Anitescu's convex
time-stepping contact model in the state transition.  For a floating-base
robot with $n_{\mathrm a}$ actuated joints, let
\begin{equation*}
\begin{aligned}
  \mathbf q_k&=({}^{\mathrm W}\mathbf X^{\mathrm B}_k,
  \mathbf q_{\mathrm a,k})\in\mathrm{SE}(3)\times\mathbb R^{n_{\mathrm a}},\\
  \mathbf v_k&=({}^{\mathrm W}\mathbf V^{\mathrm B}_{\mathrm B,k},
  \dot{\mathbf q}_{\mathrm a,k})\in\mathbb R^{6+n_{\mathrm a}},
  \qquad \mathbf x_k=(\mathbf q_k,\mathbf v_k).
\end{aligned}
\end{equation*}
The base twist is expressed in $\mathrm B$.  Let $\mathcal I_c$ index a fixed
set of candidate point contacts.  The continuous rigid-body dynamics are
\begin{equation*}
  \mathbf M(\mathbf q_k)\dot{\mathbf v}_k
  +\mathbf b(\mathbf q_k,\mathbf v_k)
  =\mathbf B\mathbf u_k
  +\sum_{i\in\mathcal I_c}
  \mathbf J_i(\mathbf q_k)^\top\mathbf f_{i,k},
\end{equation*}
where $\mathbf M$ and $\mathbf b$ are the mass matrix and bias-force vector,
$\mathbf B$ selects the actuated coordinates, $\mathbf u_k$ contains the
measured joint torques, and $\mathbf J_i$ maps generalized velocity to the
relative velocity at contact $i$, with corresponding contact force
$\mathbf f_{i,k}$.  Its normal and tangential blocks are
$\mathbf J_i=\operatorname{col}(\mathbf J_{n,i},\mathbf J_{t,i})$.
Setting the contact forces to zero over a sample period $h$ gives the
contact-free semi-implicit Euler prediction
\begin{equation*}
  \mathbf M_k(\mathbf v_{\mathrm{free},k+1}-\mathbf v_k)
  =h(\mathbf B\mathbf u_k-\mathbf b_k),
\end{equation*}
where $\mathbf M_k:=\mathbf M(\mathbf q_k)\succ\mathbf0$ and
$\mathbf b_k:=\mathbf b(\mathbf q_k,\mathbf v_k)$.

For $i\in\mathcal I_c$, let $\phi_i(\mathbf q_k)$ be the signed gap, positive
in separation, and define
\begin{equation*}
\begin{aligned}
g_i(\bar{\mathbf v})
&:=\phi_i(\mathbf q_k)/h+\mathbf J_{n,i}(\mathbf q_k)\bar{\mathbf v},\\
\mathcal K_i^*
&:=\{(a,\mathbf y)\in\mathbb R\times\mathbb R^2
\mid a\geq\mu_i\|\mathbf y\|_2\},
\end{aligned}
\end{equation*}
where $\mu_i>0$ is the friction coefficient and $\mathcal K_i^*$ is the dual
friction cone.  Anitescu's relaxation selects
the post-step velocity by minimizing
$\tfrac12\|\bar{\mathbf v}-\mathbf v_{\mathrm{free},k+1}\|_{\mathbf M_k}^2$
subject to
$(g_i(\bar{\mathbf v}),\mathbf J_{t,i}(\mathbf q_k)\bar{\mathbf v})
\in\mathcal K_i^*$ for every candidate contact~\cite{Anitescu2006}, where
$\|\mathbf a\|_{\mathbf M}^2:=\mathbf a^\top\mathbf M\mathbf a$.  The relaxation
matches Coulomb complementarity in sticking and zero-impulse separation; in
sliding it may create an artificial positive gap (spurious lift-off) whose
thickness vanishes as $h\rightarrow0$
~\cite{PangTedrake2021,PangTRO2023}.

The resulting contact map is generally only piecewise smooth.  Following the
analytic smoothing construction in~\cite{PangTRO2023}, PRIME replaces the cone
constraints by a logarithmic barrier.  Define
\begin{equation*}
\begin{aligned}
s_i(\bar{\mathbf v})
&:=\frac{g_i(\bar{\mathbf v})^2}{\mu_i^2}
-\left\|\mathbf J_{t,i}(\mathbf q_k)\bar{\mathbf v}\right\|_2^2,\\
\mathcal D_k
&:=\left\{\bar{\mathbf v}\ \middle|
g_i(\bar{\mathbf v})>
\mu_i\left\|\mathbf J_{t,i}(\mathbf q_k)\bar{\mathbf v}\right\|_2,
\ \forall i\in\mathcal I_c\right\}.
\end{aligned}
\end{equation*}
For $\kappa>0$, PRIME evaluates
\begin{equation*}
  \mathbf v_{\kappa,k+1}
  :=\arg\min_{\bar{\mathbf v}\in\mathcal D_k}
  \frac12\|\bar{\mathbf v}-\mathbf v_{\mathrm{free},k+1}\|_{\mathbf M_k}^2
  -\frac1\kappa\sum_{i\in\mathcal I_c}\log s_i(\bar{\mathbf v}).
\end{equation*}
Let $\mathbf t_i(\bar{\mathbf v}):=\mathbf J_{t,i}(\mathbf q_k)\bar{\mathbf v}$,
$\mathbf z_i(\bar{\mathbf v}):=\operatorname{col}
(g_i(\bar{\mathbf v}),\mathbf t_i(\bar{\mathbf v}))$,
$\mathcal K_i:=(\mathcal K_i^*)^*$, and
$\Delta\mathbf v_{\kappa,k}:=\mathbf v_{\kappa,k+1}
-\mathbf v_{\mathrm{free},k+1}$.  The barrier optimality condition gives
\begin{equation*}
\begin{aligned}
\boldsymbol\lambda_{\kappa,i,k}
&:=\frac{2}{\kappa s_i(\mathbf v_{\kappa,k+1})}
\begin{bmatrix}
g_i(\mathbf v_{\kappa,k+1})/\mu_i^2\\
-\mathbf t_i(\mathbf v_{\kappa,k+1})
\end{bmatrix},\\
\mathbf M_k\Delta\mathbf v_{\kappa,k}
&=\sum_{i\in\mathcal I_c}\mathbf J_i(\mathbf q_k)^\top
\boldsymbol\lambda_{\kappa,i,k},\\
\left\langle\boldsymbol\lambda_{\kappa,i,k},
\mathbf z_i(\mathbf v_{\kappa,k+1})\right\rangle&=\frac{2}{\kappa}.
\end{aligned}
\end{equation*}
Here $\boldsymbol\lambda_{\kappa,i,k}\in\operatorname{int}\mathcal K_i$ and
$\mathbf z_i(\mathbf v_{\kappa,k+1})\in\operatorname{int}\mathcal K_i^*$.
The unsmoothed KKT system retains the impulse balance but replaces interior
membership and $2/\kappa$ by closed-cone membership and zero pairing.  Thus
$2/\kappa$ is the per-contact perturbed pairing, and
$\boldsymbol\lambda_{\kappa,i,k}/h$ is the step-average force
~\cite{Kang2026PRIME}.

If $\mathcal D_k$ is nonempty, $\mathbf M_k\succ\mathbf0$ makes the
finite-$\kappa$ velocity Hessian positive definite.  The minimizer is therefore
unique and converges to the unsmoothed solution as $\kappa\rightarrow\infty$
~\cite{PangTRO2023}.  For locally smooth problem data and fixed smooth contact
geometry, the implicit function theorem supplies local derivatives with respect
to the state, torque, and geometry.  Configuration integration then defines the
locally smooth process map
$\mathbf F_{\kappa}(\mathbf x_k,\mathbf u_k;\boldsymbol\rho)$, with
$\boldsymbol\rho$ entering through the signed gaps, contact frames, and contact
Jacobians.  Contact modes need not be prescribed, and impulses remain latent.
Process uncertainty is represented by the stochastic residual in
Section~\ref{sec:map}.

\subsubsection{Measurement Modeling}\label{sec:prime_measurement}

The measurement vector contains the observed encoder joint positions and
velocities together with the available pose and velocity channels of the
motion-capture body $\mathrm M$.  Joint channels are selected from
$(\mathbf q_k,\mathbf v_k)$, whereas the motion-capture prediction applies the
base-to-marker offset $\boldsymbol\rho_{\mathrm{BM}}$ defined in
Section~\ref{sec:problem}.  Only observed channels enter the residual, whose
additive zero-mean Gaussian noise has covariance $\mathbf R_k$.  The measured
joint torque is treated as the known process input $\mathbf u_k$ and is
therefore not assigned a separate measurement residual.  No contact labels or
contact-force measurements are required; their dynamical effects are
represented by the smoothed contact step.

\subsection{Optimization-Based Estimation}\label{sec:map}

Under the first-order Markov assumption and independent Gaussian process and
measurement disturbances, FIE combines the initial-state prior with the
negative log-likelihood of the complete input--measurement record.  Up to terms
independent of the lower-level variables, the resulting problem is
\begin{align}
\min_{\mathcal Z_{0:T}}\quad&
  \Gamma(\mathbf x_0)
  +\sum_{k=0}^{T-1}\|\mathbf w_k\|_{\mathbf Q_k^{-1}}^2
  +\sum_{k=0}^{T}\|\mathbf n_k\|_{\mathbf R_k^{-1}}^2
  \notag\\
\mathrm{s.t.}\quad&
  \mathbf r^x_k(\mathbf x_k,\mathbf x_{k+1},
  \mathbf u_k;\boldsymbol\rho)=\mathbf w_k,
  \quad k=0,\ldots,T-1,\notag\\
& \mathbf y_k
  =\mathbf H(\mathbf x_k,\mathbf u_k;\boldsymbol\rho)+\mathbf n_k,
  \quad k=0,\ldots,T.
\label{eq:FIE_formulation}
\end{align}
Here $\mathbf x_{0:T}:=\{\mathbf x_k\}_{k=0}^{T}$ and
$\mathcal Z_{0:T}:=(\mathbf x_{0:T},\mathbf w_{0:T-1},
\mathbf n_{0:T})$.  The function $\Gamma$ is the negative log prior on the initial state,
and $\mathbf r^x_k$ and $\mathbf H$ denote the selected process residual and
measurement model.  The disturbances satisfy
$\mathbf w_k\sim\mathcal N(\mathbf0,\mathbf Q_k)$ and
$\mathbf n_k\sim\mathcal N(\mathbf0,\mathbf R_k)$, with
$\|\mathbf a\|_{\boldsymbol\Sigma^{-1}}^2
:=\mathbf a^\top\boldsymbol\Sigma^{-1}\mathbf a$.

Instantiating $\mathbf r^x_k$ and $\mathbf H$ with the models in
\hyperref[sec:mobile]{Section~\ref*{sec:mobile}} gives the \emph{fast FIE},
whereas using the smoothed contact process and measurements in
\hyperref[sec:prime_estimator]{Section~\ref*{sec:prime_estimator}} gives the
\emph{PRIME FIE}.  Hereafter, \emph{FIE} denotes the selected formulation.

\section{Problem Statement}\label{sec:problem}

We jointly calibrate the process and measurement covariances and uncertain
geometry of the FIE from a training trajectory with ground truth.  The geometry
affects the kinematic observations and, in the contact-aware formulation, the
process model.

\subsection{Data and Calibration Objective}

\block{Training data}
The robot carries an IMU and joint encoders and records the inputs required by
the selected formulation.  During calibration, an external motion-
capture system measures the pose and velocity of a marker body $\mathrm M$
rigidly attached to the torso.  For a trajectory of length $T$, let
$\mathbf U:=\{\mathbf u_k\}_{k=0}^{T-1}$ and
$\mathbf Y:=\{\mathbf y_k\}_{k=0}^{T}$ denote the recorded inputs and sensor
data, respectively, and let
$\mathbf x_{\mathrm{GT},0:T}:=\{\mathbf x_{\mathrm{GT},k}\}_{k=0}^{T}$ collect
the available ground-truth channels.  Ground truth need not cover latent states
such as IMU biases.  We assume that the training motion excites the calibrated
parameters sufficiently; a formal identifiability analysis is outside the
present scope.

\block{Calibration variables}
We separate statistical and geometric parameters as
\begin{equation}
  \boldsymbol\theta=(\boldsymbol\alpha,\boldsymbol\rho),
  \qquad
  \boldsymbol\rho=(\boldsymbol\rho_{\mathrm{foot}},
  \boldsymbol\rho_{\mathrm{BM}}).
  \label{eq:calibration_parameters}
\end{equation}
The coordinates $\boldsymbol\alpha$ parameterize all calibrated process and
measurement covariance blocks.  The vector $\boldsymbol\rho_{\mathrm{foot}}$
collects the uncertain terminal-link geometry, while
$\boldsymbol\rho_{\mathrm{BM}}:={}^{\mathrm B}\mathbf p^{\mathrm M}_{\mathrm B}
\in\mathbb R^3$ is the position of the motion-capture body origin measured
from the floating-base origin and expressed in $\mathrm B$.  The latter is
needed to compare the estimated base trajectory with marker-based ground
truth.

For leg $j$, let $\mathrm L_j$ be the terminal-link frame and $\mathrm P_j$
the physical foot point represented by the geometry correction.  We define
\begin{equation*}
  \boldsymbol\rho_{\mathrm{foot},j}
  :={}^{\mathrm F_j}\mathbf p^{\mathrm P_j}_{\mathrm L_j}\in\mathbb R^3.
\end{equation*}
Thus the corrected point in the world frame is
\begin{equation*}
  {}^{\mathrm W}\mathbf p^{\mathrm P_j}_{\mathrm W}
  ={}^{\mathrm W}\mathbf p^{\mathrm F_j}_{\mathrm W}
  +{}^{\mathrm W}\mathbf R^{\mathrm L_j}
  \boldsymbol\rho_{\mathrm{foot},j}.
\end{equation*}
This endpoint correction represents aggregate distal-link error; richer
geometric parameterizations may be introduced when supported by the data.  In
the contact-conditioned kinematic formulation,
$\boldsymbol\rho_{\mathrm{foot}}$ affects the observations in
\eqref{eq:meas_models} and their propagated covariances through
\eqref{eq:kinematic-residual-jacobian}.  In the contact-aware formulation, it
additionally affects the
signed gaps, contact frames, Jacobians, and process map.

\subsection{Bi-Level Calibration}

Using the state-trajectory form of the FIE in Section~\ref{sec:map}, obtained
by eliminating the disturbance variables through its equality constraints,
let $\mathcal X$ denote the feasible trajectory set and $\mathcal J$ the
resulting lower objective.  The calibration problem is
\begin{align*}
  \min_{\boldsymbol\theta\in\mathbb R^{n_\theta}}\quad&
  \mathcal L(\boldsymbol\theta)\\
  \mathrm{s.t.}\quad&\boldsymbol\theta\in\mathcal C,\\
  &\widehat{\mathbf x}_{0:T}(\boldsymbol\theta)
  :=\{\widehat{\mathbf x}_k(\boldsymbol\theta)\}_{k=0}^{T}
  \in\arg\min_{\mathbf x_{0:T}\in\mathcal X}
  \mathcal J(\mathbf x_{0:T};\boldsymbol\theta).
\end{align*}
The reduced loss compares the parameter-dependent estimate
$\widehat{\mathbf x}_{0:T}(\boldsymbol\theta)$ with the available reference
channels after the geometry-dependent marker-to-base transformation.  The
statistical coordinates $\boldsymbol\alpha$ determine the lower-level
information matrices, while the geometry $\boldsymbol\rho$ enters the relevant
process, measurement, and reference transformations.

\subsection{Feasible Calibration Set}

For covariance blocks
$\{\boldsymbol\Sigma_i\in\mathbb R^{d_i\times d_i}\}_{i=1}^{n_\Sigma}$ and
fixed margins $\epsilon_i>0$, the
corresponding entries of $\boldsymbol\alpha$ stack their upper triangles in a
fixed order.  We restrict the complete calibration vector to
\begin{equation}
\mathcal C:=\left\{
  \boldsymbol\theta\in\mathbb R^{n_\theta}\ \middle|\
  \begin{aligned}
  \underline{\boldsymbol\theta}&\leq\boldsymbol\theta
  \leq\overline{\boldsymbol\theta},\\
  \boldsymbol\Sigma_i(\boldsymbol\alpha)&\succeq\epsilon_i\mathbf I,
  \quad i=1,\ldots,n_\Sigma
  \end{aligned}
\right\}.
\label{eq:calibration_set}
\end{equation}
The elementwise bounds encode available sensor and geometry knowledge.  The
fixed eigenvalue margins keep the covariance blocks uniformly positive
definite and prevent their information matrices from becoming singular.

Since each $\boldsymbol\Sigma_i$ is affine in $\boldsymbol\alpha$, the nonempty
set $\mathcal C$ is a compact convex spectrahedron.

\section{Method}\label{sec:method}

We differentiate the lower FIE solution with respect to the calibration
parameters and use the resulting first-order oracle in three constrained
upper-level methods.  Let $\boldsymbol{\chi}$ collect the primal variables of
\eqref{eq:FIE_formulation}, let $\boldsymbol{\nu}$ collect its
equality-constraint multipliers, and write
$\mathbf{z}^{\star}(\boldsymbol{\theta})
=(\boldsymbol{\chi}^{\star},\boldsymbol{\nu}^{\star})$ for a lower
primal--dual solution.  The estimated trajectory
$\widehat{\mathbf x}_{0:T}(\boldsymbol\theta)$ is selected from
$\boldsymbol{\chi}^{\star}$.  The symbol $k$ denotes estimator time and $t$
denotes an upper-level iteration.

\subsection{Upper-Level Loss and First-Order Gradient}
\label{sec:upper-loss-gradient}

\subsubsection{Trajectory loss}\label{sec:costgrad}

Write the estimated stage state as
$\widehat{\mathbf x}_k=({}^{\mathrm W}\widehat{\mathbf X}^{\mathrm B}_k,
\widehat{\mathbf x}_{\mathrm E,k})\in
\mathrm{SE}(3)\times\mathbb R^d$, where
${}^{\mathrm W}\widehat{\mathbf X}^{\mathrm B}_k$ is the floating-base pose
defined in~\eqref{eq:state_def} and
$\widehat{\mathbf x}_{\mathrm E,k}$ collects the Euclidean components of the
selected estimator.  Let
$\widehat{\mathbf x}_{\mathrm o,k}(\boldsymbol\theta)$ and
$\mathbf x_{\mathrm{GT},\mathrm o,k}(\boldsymbol\theta)$ denote those Euclidean
components for which reference values are available, expressed in the
corresponding frames.
The reference pose
${}^{\mathrm W}\mathbf X^{\mathrm B}_{\mathrm{GT},k}
(\boldsymbol\theta)$ follows the same frame convention and incorporates the
calibrated base-to-marker position offset.
The trajectory residual is
\begin{equation}
\mathbf{e}_k(\boldsymbol{\theta})
=
\begin{bmatrix}
\operatorname{Log}\!\left(
\left({}^{\mathrm W}\mathbf X^{\mathrm B}_{\mathrm{GT},k}
(\boldsymbol\theta)\right)^{-1}
{}^{\mathrm W}\widehat{\mathbf X}^{\mathrm B}_k(\boldsymbol\theta)
\right)^{\!\vee}\\
\widehat{\mathbf x}_{\mathrm o,k}(\boldsymbol\theta)
-\mathbf x_{\mathrm{GT},\mathrm o,k}(\boldsymbol\theta)
\end{bmatrix}.
\label{eq:trajectory_error}
\end{equation}
Here $\operatorname{Log}:\mathrm{SE}(3)\rightarrow\mathfrak{se}(3)$ is the
local inverse of $\operatorname{Exp}$ on the principal branch, restricted to
relative poses whose rotational part has angle strictly below $\pi$; the vee
operator returns fixed vector coordinates with translation preceding
rotation.  State components without reference values do not enter the upper
objective.  The pose error in
\eqref{eq:trajectory_error} is unchanged by a common change of the world frame
and avoids a coordinate subtraction of rotation
parameters~\cite{barfoot2024state,teng2022liecost}.

The upper objective is the weighted trajectory discrepancy
\begin{equation*}
\mathcal{L}(\boldsymbol{\theta})
=\frac{1}{2}\sum_{k=0}^{T}
  \mathbf{e}_k(\boldsymbol{\theta})^{\top}
  \mathbf{W}_k\mathbf{e}_k(\boldsymbol{\theta}),
\qquad \mathbf{W}_k\succeq0.
\end{equation*}
The fixed matrices $\mathbf W_k$ set the relative scale of translational,
rotational, and Euclidean channels and are independent of the estimator
covariances calibrated in the lower level.

\subsubsection{First-order gradient}\label{sec:grad_upper}

The gradient of the reduced objective follows from the KKT
system of the parameterized FIE in~\eqref{eq:FIE_formulation}.  Let
$\mathbf c(\boldsymbol\chi;\boldsymbol\theta)=\mathbf0$ collect its equality
constraints, and define the lower Lagrangian
$\mathscr J(\boldsymbol\chi,\boldsymbol\nu;\boldsymbol\theta)
:=\mathcal J(\boldsymbol\chi;\boldsymbol\theta)
+\boldsymbol\nu^{\top}\mathbf c(\boldsymbol\chi;\boldsymbol\theta)$.  In local
coordinates, the primal--dual residual is
\begin{equation*}
\mathcal{F}(\mathbf{z},\boldsymbol{\theta})
=\begin{bmatrix}
\nabla_{\boldsymbol{\chi}}\mathscr{J}
 (\boldsymbol{\chi},\boldsymbol{\nu};\boldsymbol{\theta})\\
\mathbf{c}(\boldsymbol{\chi};\boldsymbol{\theta})
\end{bmatrix},
\qquad
\mathcal{F}(\mathbf{z}^{\star},\boldsymbol{\theta})=\mathbf{0}.
\end{equation*}
All derivatives below are evaluated at
$(\mathbf z^\star(\boldsymbol\theta),\boldsymbol\theta)$.  Define
\begin{equation*}
\mathbf{K}:=\frac{\partial\mathcal{F}}{\partial\mathbf{z}},
\qquad
\mathbf{G}:=\frac{\partial\mathcal{F}}
                  {\partial\boldsymbol{\theta}}.
\end{equation*}
If $\mathbf K$ is nonsingular, differentiating
$\mathcal F(\mathbf z^\star(\boldsymbol\theta),\boldsymbol\theta)=\mathbf0$
with respect to $\boldsymbol\theta$ yields
\begin{equation*}
\mathbf{K}
\frac{D\mathbf{z}^{\star}}{D\boldsymbol{\theta}}
=-\mathbf{G}.
\end{equation*}
The upper-loss differential with respect to $\mathbf z$, with zero components
for variables absent from the trajectory loss, forms the right-hand side of
the transposed KKT system defining the adjoint $\boldsymbol\lambda$.  The
first-order gradient combines the explicit partial derivative of the upper
loss with the implicit pullback $-\mathbf G^{\top}\boldsymbol\lambda$.  The
explicit term accounts for the dependence of the ground-truth transformation
on the geometric parameters, whereas $\mathbf G$ collects the dependence of
the lower KKT system on all calibration parameters through the covariance
weights and residual models.
The resulting upper-level interface is the first-order oracle
\begin{equation}
\mathcal{O}(\boldsymbol{\theta})
=\bigl(\mathcal{L}(\boldsymbol{\theta}),
       \nabla_{\boldsymbol{\theta}}\mathcal{L}(\boldsymbol{\theta})\bigr).
\label{eq:calibration-oracle}
\end{equation}

\subsection{Upper-Level Updates}\label{sec:upper-methods}

The oracle~\eqref{eq:calibration-oracle} supplies the reduced loss and its
gradient, but not a reduced Hessian.  Upper-level updates must also
respect the coupled positive-definite constraints in
\eqref{eq:calibration_set}; in particular, entrywise clipping of covariance
parameters does not preserve feasibility.  We therefore consider three
alternative strategies---Frank--Wolfe with a semidefinite linear minimization
oracle (LMO), SQP with BFGS curvature, and projected Adam.  All three use the
same first-order oracle and differ only in how they construct a step and
maintain feasibility.  The bi-level formulation does not privilege any of
these update rules.

Frank--Wolfe acts directly on $\boldsymbol{\theta}$.  For SQP and projected
Adam, we introduce scaled coordinates
$\boldsymbol\eta\in[0,1]^{n_\eta}$ and decode
each covariance block according to
\begin{equation*}
\boldsymbol{\Sigma}_i(\boldsymbol\eta)
=\epsilon_i\mathbf{I}+\mathbf{L}_i(\boldsymbol\eta)
 \mathbf{L}_i(\boldsymbol\eta)^{\top},
\end{equation*}
where $\mathbf{L}_i$ is lower triangular with positive diagonal.  Its free
entries, together with the geometric parameters, define a smooth decoding
$\boldsymbol{\theta}=\boldsymbol{\psi}(\boldsymbol\eta)$ and enforce
$\boldsymbol{\Sigma}_i\succeq\epsilon_i\mathbf I$ at every trial point.
Remaining admissibility conditions are written as
$\mathbf c_{\mathrm{up}}(\boldsymbol\eta)\leq\mathbf0$.  Gradients in these
coordinates follow from the chain rule,
$\nabla_{\boldsymbol\eta}(\mathcal L\circ\boldsymbol\psi)
=D\boldsymbol\psi^{\top}\nabla_{\boldsymbol\theta}\mathcal L$.

\subsubsection{Frank--Wolfe with a semidefinite oracle}

Frank--Wolfe preserves the original covariance representation and minimizes
the first-order model of the reduced loss over the calibration set.  The LMO
is the following semidefinite program, where
$\mathbf{s}=(\mathbf{s}_{\alpha},\mathbf{s}_{\rho})$ follows the partition
in~\eqref{eq:calibration_parameters}:
\begin{align*}
\mathbf{s}^{t}
=\underset{\mathbf{s}}{\operatorname{argmin}}\quad
& \bigl(\nabla_{\boldsymbol{\theta}}\mathcal{L}
  (\boldsymbol\theta^t)\bigr)^{\top}\mathbf{s}\\
\text{subject to}\quad
& \underline{\boldsymbol{\theta}}^{t}\leq\mathbf{s}
  \leq\overline{\boldsymbol{\theta}}^{t},\\
& \boldsymbol{\Sigma}_i(\mathbf{s}_{\alpha})-\epsilon_i\mathbf{I}
  \succeq\mathbf{0},\qquad i=1,\ldots,n_{\Sigma}.
\end{align*}
Here $[\underline{\boldsymbol\theta}^{t},
\overline{\boldsymbol\theta}^{t}]$ is either the global box or its intersection
with a box trust region centered at $\boldsymbol{\theta}^{t}$.  Each covariance
block is affine in $\mathbf{s}_{\alpha}$, so the oracle has a linear objective
and linear matrix inequalities.  The oracle therefore retains the coupled
covariance geometry without a projection onto the semidefinite set.

Because $\boldsymbol\theta^t$ and $\mathbf s^t$ belong to the convex set
$\mathcal C$, the update
$\boldsymbol\theta^t+\gamma_t(\mathbf s^t-\boldsymbol\theta^t)$ remains in
$\mathcal C$ for every $\gamma_t\in[0,1]$.  A backtracking or diminishing step
rule determines the accepted convex combination.  The Frank--Wolfe gap
$\langle\nabla_{\boldsymbol\theta}\mathcal L(\boldsymbol\theta^t),
\boldsymbol\theta^t-\mathbf s^t\rangle$ provides the associated first-order
stationarity measure~\cite{pmlr-v28-jaggi13}.

\subsubsection{Sequential quadratic programming with BFGS curvature}

SQP constructs a constrained quadratic model in the scaled Cholesky
coordinates.  At iteration $t$, the step is obtained from
\begin{align*}
\underset{\mathbf{p}}{\operatorname{minimize}}\quad
& \frac{1}{2}\mathbf{p}^{\top}\mathbf{B}^{t}\mathbf{p}
  +(\mathbf{g}^{t})^{\top}\mathbf{p}\\
\text{subject to}\quad
& \mathbf c_{\mathrm{up}}(\boldsymbol\eta^{t})
  +\mathbf J_{\mathrm{up}}(\boldsymbol\eta^{t})\mathbf{p}
  \leq\mathbf{0},\\
& -\boldsymbol\eta^{t}\leq\mathbf{p}\leq\mathbf{1}-\boldsymbol\eta^{t}.
\end{align*}
where $\mathbf g^t:=\nabla_{\boldsymbol\eta}
(\mathcal L\circ\boldsymbol\psi)(\boldsymbol\eta^t)$,
$\mathbf J_{\mathrm{up}}$ is the constraint Jacobian, and
$\mathbf B^t$ approximates the Hessian of the upper-level Lagrangian.  The
matrix $\mathbf B^t$ is initialized by a scaled identity and updated by damped
BFGS from successive Lagrangian gradients, avoiding second-order
differentiation of the lower solution map.  A standard merit-function line
search globalizes the step~\cite{nocedal2006numerical}.

\subsubsection{Projected Adam}

Projected Adam is used when the admissible coordinate set admits a tractable
Euclidean projection.  It applies the standard bias-corrected first- and
second-moment recursions to
$\nabla_{\boldsymbol\eta}(\mathcal L\circ\boldsymbol\psi)$ and projects each
tentative step onto
$\mathcal E=\{\boldsymbol\eta\in[0,1]^{n_\eta}:
\mathbf c_{\mathrm{up}}(\boldsymbol\eta)\leq\mathbf0\}$.
The update uses neither a quadratic subproblem nor a line search and forms its
diagonal rescaling from the gradient history~\cite{kingma2015adam}.

\subsection{Numerical Acceleration}\label{sec:method-implementation}

Repeated evaluations of the common oracle are accelerated in three ways.

\subsubsection{Adjoint factorization and reuse}

Because the upper objective is scalar, its gradient requires one transposed
KKT solve after each accepted lower solution.  The equality-constrained FIE
has a symmetric KKT linearization; a fill-reducing permutation gives

\begin{equation*}
\mathbf P\mathbf K\mathbf P^{\top}
 =\mathbf L\mathbf D\mathbf L^{\top}.
\end{equation*}
The factors are used for the adjoint solve and reused while $\mathbf K$ remains
unchanged; a new calibration point generally requires a new numerical
factorization~\cite{boyd2011admm}.

\subsubsection{OCP structure and setup amortization}

The FIE variables and adjacent-stage process constraints are ordered by time
and passed to Fatrop with explicit stage dimensions, allowing its generalized
Riccati recursion to exploit the resulting optimal-control
structure~\cite{vanroye2023fatrop}.  The symbolic OCP
and its derivative functions are constructed once per horizon with
CasADi~\cite{Andersson2019} and reused across upper iterations.  Fatrop solves
the lower OCP; the sparse factors above are used only for the adjoint gradient
evaluation.

\subsubsection{Warm starts and work control}

Consecutive lower problems are initialized from the preceding accepted primal
solution; a failed warm solve is retried from a model-derived cold
initialization.  A lower solution is accepted when
\begin{equation*}
\|\mathcal{F}(\mathbf{z}^{\star},\boldsymbol{\theta})\|_{\infty}
\leq\epsilon_{\mathrm{kkt}},
\qquad
\|\mathbf{c}(\boldsymbol{\chi}^{\star};
             \boldsymbol{\theta})\|_{\infty}
\leq\epsilon_{\mathrm{feas}}.
\end{equation*}
A failed lower trial causes a backtracking upper method to reduce its step.
Upper termination uses the SQP optimality residual, the Frank--Wolfe gap, or
the prescribed Adam budget, together with feasibility and step tests.

\section{Results}\label{sec:results}
We evaluate our calibration approach on three robots: a bipedal robot STRIDE \cite{Stride} and two quadrupedal robot Go1 and B1 from Unitree. The data used from STRIDE and Go1 are obtained from simulation in Matlab and MuJoCo environment in their open source repository \cite{Stride}. We obtained the data of B1 on hardware in the motion capture room with 12 Opti-track Cameras, with a combination of Prime 13 and 22. The bi-level optimization is implemented in Python and C++. The outer-loop LMO is solved by MOSEK, and the lower-level FIE is solved by Fatrop~\cite{vanroye2023fatrop}. The analytical gradients are generated via CasADi~\cite{Andersson2019} and Pinocchio \cite{carpentier:hal-01866228}.

\begin{figure}[!htbp]
    \centering
    \includesvg[width=0.9\linewidth]{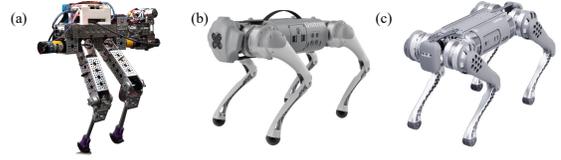}
        \caption{The robot STRIDE (a), quadrupedal robot Go1 (b), and B1 (c) are used in the evaluation (Pictures are used with permission). }
    \label{fig:robots}
\end{figure}

We report the upper-level optimization behavior and the resulting covariance
and geometric calibration across the three platforms.  The experiments cover
different morphologies, dynamics models, and sensing conditions; the platforms
are shown in Fig.~\ref{fig:robots}.

\block{Upper-Level Update Comparison}

We compare the three upper-level update schemes under a common initialization, trajectory loss, and first-order oracle for each FIE formulation. As shown in Figure~\ref{fig:upper-optimizer-comparison}, SQP--BFGS and Frank--Wolfe with a semidefinite oracle achieve comparable final losses. Projected Adam exhibits less monotone progress and greater variability across iterations.

\begin{figure}[!htbp]
    \centering
    \includesvg[width=0.94\columnwidth]{figure/upper_optimizer_comparison.svg}
    \caption{Convergence of the upper-level loss for (a) the fast FIE and
    (b) the PRIME FIE. Methods share the initialization within each panel,
    and rejected trials are omitted. Loss scales are formulation-specific.}
    \label{fig:upper-optimizer-comparison}
\end{figure}

\begin{figure}[!htbp]
    \centering
    \includesvg[width=0.9\linewidth]{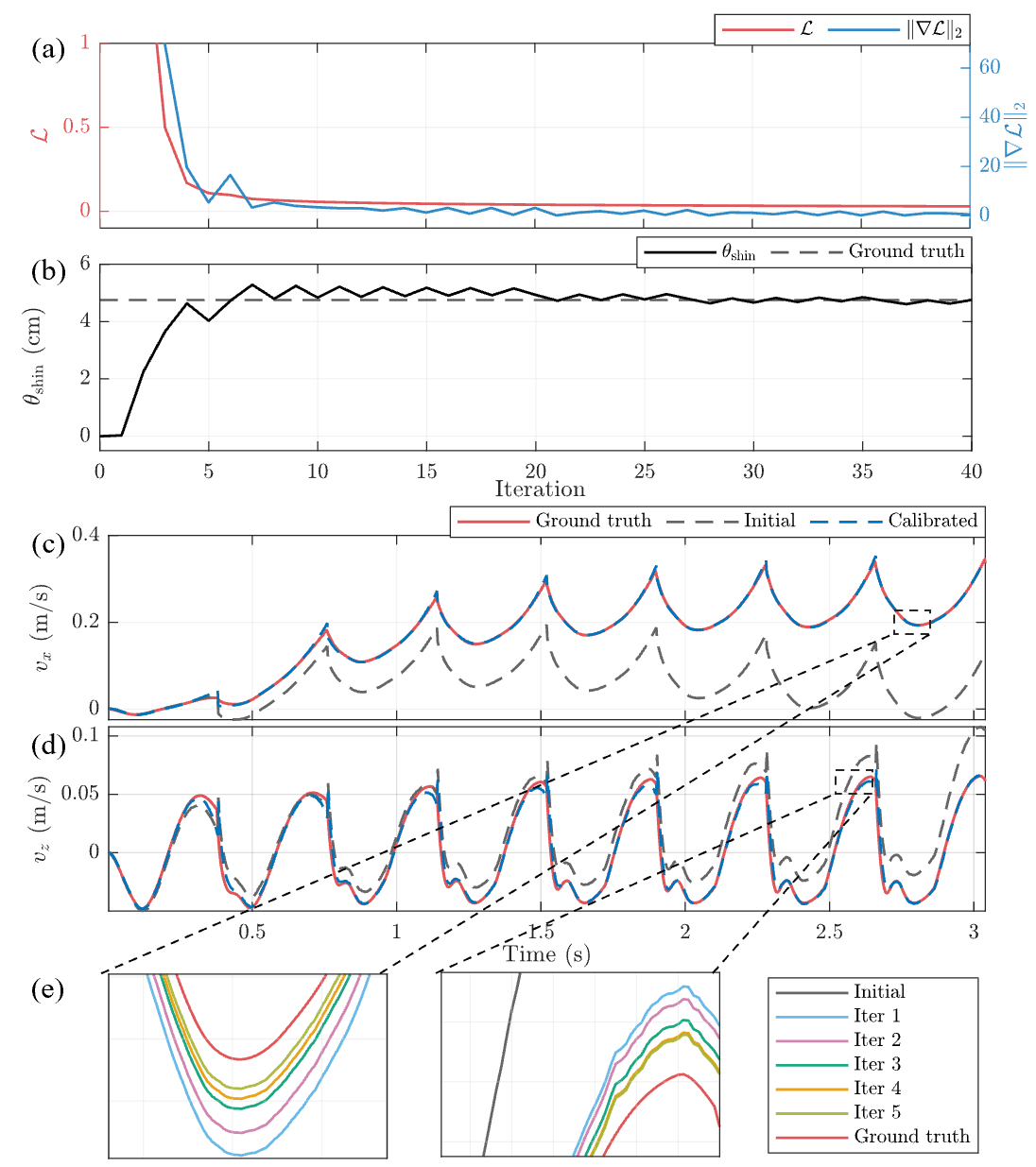}
 \caption{Calibration results on STIRDE: (a) the convergence of the upper loss function and norm of the gradients, (b) the convergence of kinematics bias w.r.t. ground truth, (c) and (d) the linear velocities of the ground truth, initial estimates, and calibrated estimates. }
    \label{fig:STRIDE}
\end{figure}

\block{Joint Sensor Noise \& Kinematics Calibration} For the robot STRIDE, we mainly evaluate the joint calibration on the sensor noises and robot kinematics since their dynamics processes are deterministic in Matlab. \fix{We injected additive white Gaussian noise into the joint encoder angle/velocity measurements, and corrupt the IMU with non-diagonal correlated Gaussian noise, along with an articulated shin length offset to the estimator.} Fig. \ref{fig:STRIDE} shows the calibration results. The loss successfully reduced through the iterations of the bi-level optimization, and estimated velocity converges to the ground truth; the kinematic parameters converge as well.

\begin{figure}[!htbp]
    \centering
    \includesvg[width=0.9\linewidth]{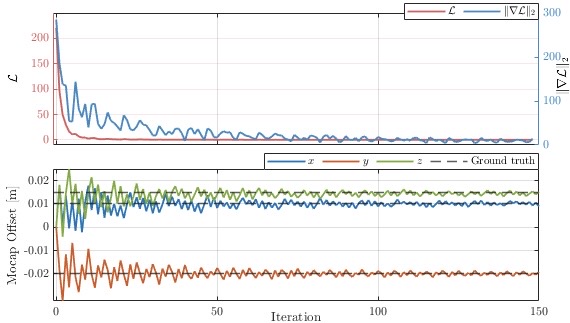}
    \caption{Calibration results on Go1: convergences of cost and gradient (top), and kinematic offset (bottom).}
    \label{fig:go1}
\end{figure}

\block{Joint Process \& Sensor Noise Calibration}
For the robot Go1, we evaluate the calibration using data generated by MuJoCo’s time-stepping, multi-contact simulation. Sensor noise is defined in the MJCF/XML and evaluated at every step as a function of the simulated state. Figure~\ref{fig:go1} shows the reduction in loss and gradient norm and the convergence of the injected motion-capture offset toward its prescribed value.

\begin{figure}[!htbp]
    \centering
    \includesvg[width=0.98\linewidth]{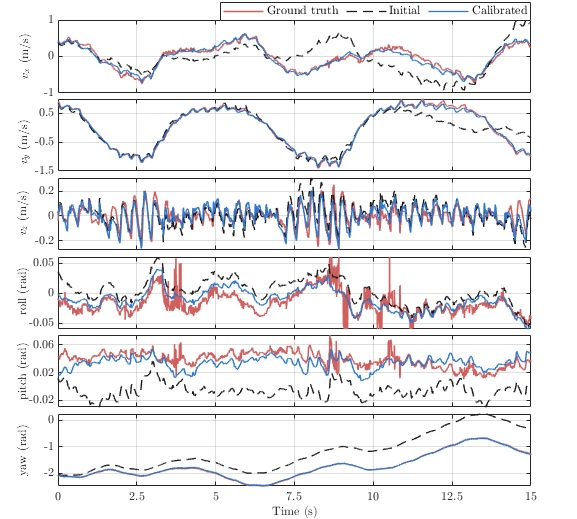}
    \caption{Calibration results on B1 hardware in terms of linear velocity and torso orientation.}
    \label{fig:B1estimate}
\end{figure}

\begin{figure}[!htbp]
    \centering
    \includesvg[width=0.88\linewidth]{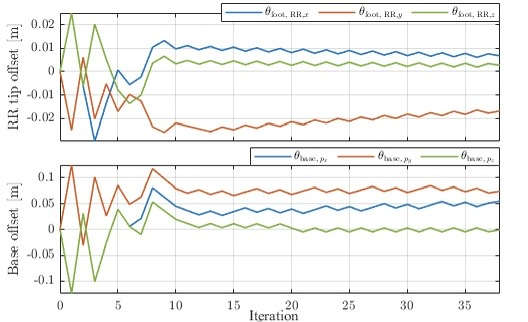}
    \caption{Convergence of the kinematics calibration on B1 robot
    hardware: (top) the rear-right foot offset and (bottom) the
    base-to-marker offset.}
    \label{fig:B1kinematics}
\end{figure}

\begin{figure}[!htbp]
    \centering
    \includesvg[width=0.98\linewidth]{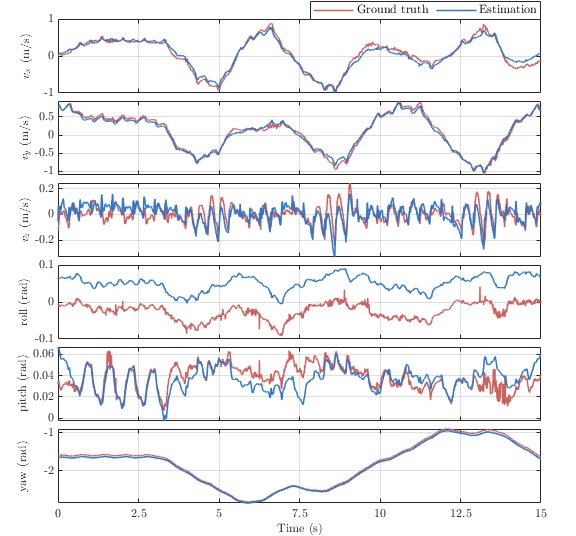}
    \caption{Evaluation results on a held-out B1 hardware segment in
    terms of linear velocity and torso orientation.}
    \label{fig:B1_evaluation}
\end{figure}

\block{Joint Noise \& Kinematics Calibration} Last and most importantly, we calibrate the robot B1 using hardware data. The robot is controlled to walk in the motion-capture room. The torso motion is measured by the motion capture system, taken as ground truth with a constant unknown offset to the base frame position of the torso. The orientation measurement of the torso is assumed to be unbiased, because the torso and the marked rigid body are well-aligned in the beginning of the experiment. Additionally, we assume all the shins of B1 have kinematic errors from the factory URDF. After we obtained the sensory data along with the ground truth measurements, we applied our calibration to B1. Fig. \ref{fig:B1estimate} shows the improved estimates of the torso's orientation and linear velocity, and Fig. \ref{fig:B1kinematics} shows the kinematics converge to reasonable values: in our experimental setup, the marked torso location is indeed about centimeters away from the base frame location. Table \ref{tab:B1_RMSE} shows the quantitative results in terms of RMSE; Fig. \ref{fig:B1_evaluation} shows the new evaluation on a different segment of data using calibrated covariances and kinematics, which still yields highly accurate estimates. 

\block{Computational Performance}
With the numerical accelerations in Section~\ref{sec:method-implementation},
wall-clock calibration time on an Intel Core Ultra~9 275HX CPU, including all
lower-level solves, averaged $27.8$\,s over four fast FIE trials and $103.7$\,s across the three PRIME FIE trials on a
$30$\,s trajectory ($3000$ samples). One-time symbolic
construction was excluded. 

For long-sequence EKF/InEKF calibration, we also provide a
\href{https://github.com/DLinC3/LegBiCal/tree/main/cuda}{CUDA/PyTorch
implementation}. The GPU executes the fixed-shape, batched filter replay and
its reverse-mode differentiation through propagation, correction, covariance
updates, and chunked backpropagation through time. On an NVIDIA RTX~5090
Laptop GPU, the complete $20$-epoch calibration of approximately $26$\,min of
trajectory data ($1.47$ million supervised time steps) takes $228.4$\,s.

 \begin{table}[!htbp]
    \centering
    \caption{RMSEs of the estimates on B1 robot hardware.}
    \small
    \begin{tabular}{@{}lccc@{}}
        \toprule
        Metric & Before & After & New evaluation \\
        \midrule
        $\mathrm{RMSE}_v$ [m/s] & 0.2658 & \textbf{0.0610} & \textbf{0.0706} \\
        $\mathrm{RMSE}_{\mathrm{Euler}}$ [rad] & 0.3457 & \textbf{0.0151} & \textbf{0.0514} \\
        \bottomrule
    \end{tabular}
    \label{tab:B1_RMSE}
\end{table}

\section{Conclusion and Future Work}
We presented a bi-level optimization framework for jointly calibrating noise covariances and kinematics in legged robot state estimation. By differentiating through a full-information estimator, our method removes manual tuning and improves accuracy and consistency across simulated and real robots. Future work will focus on reducing computational cost, analyzing observability, and extending the approach to dynamics and inertia identification for fully self-calibrating systems. 



\bibliographystyle{IEEEtran}
\bibliography{reference}
\end{document}